\documentclass{article}

\PassOptionsToPackage{numbers, compress}{natbib}
\usepackage[final]{nips_2017}

\usepackage[utf8]{inputenc} 
\usepackage[T1]{fontenc}    
\usepackage{hyperref}       
\usepackage{url}            
\usepackage{booktabs}       
\usepackage{amsfonts}       
\usepackage{nicefrac}       
\usepackage{microtype}      
\usepackage{graphicx}
\usepackage{amsmath,amssymb}
\usepackage{enumitem}
\usepackage{tabularx,ragged2e,booktabs,caption}

\title{Hybrid Gradient Boosting Trees and Neural Networks for Forecasting Operating Room Data}

\author{Hugh Chen\\
University of Washington\\
\texttt{\{hughchen\}@uw.edu} \\
\And
Scott Lundberg  \\
University of Washington \\
\texttt{\{slund1\}@cs.washington.edu}
\And 
Su-In Lee \\
University of Washington \\
\texttt{\{suinlee\}@cs.washington.edu}
}

\begin{document}

\maketitle

\begin{abstract}
Time series data constitutes a distinct and growing problem in machine learning.  As the corpus of time series data grows larger, deep models that simultaneously learn features and classify with these features can be intractable or suboptimal.  In this paper, we present feature learning via long short term memory (LSTM) networks and prediction via gradient boosting trees (XGB).   Focusing on the consequential setting of electronic health record data, we predict the occurrence of hypoxemia five minutes into the future based on past features. We make two observations: 1)  long short term memory networks are effective at capturing long term dependencies based on a single feature and 2) gradient boosting trees are capable of tractably combining a large number of features including static features like height and weight. With these observations in mind, we generate features by performing "supervised" representation learning with LSTM networks. Augmenting the original XGB model with these  features gives significantly better performance than either individual method.
\end{abstract}

\section{Background}
\label{background}
In hospitals, data is constantly recorded -- often in the form of physiological signals such as blood oxygen, heart rate, blood pressure and more.  In this paper, we address the problem of hypoxemia (low arterial blood oxygen) within operating rooms; one of the common yet potentially serious concerns that anesthesiologists deal with while safeguarding patients. In particular, one recent study found that hypoxemic events occur every 29 surgery hours on average for two hospitals’ operating rooms \citep{efvmsg10}.  These events can cause serious patient harm during general anesthesia and surgery and are correlated with cardiac arrest, cardiac arryhythmias, decreased cognitive function, and more \citep{dbch14, sn01}.

In order to assist anesthesiologists, Lundberg et. al. (2017) developed a method for hypoxemia prediction on  operating room data with the aim of interpretability \citep{Lundberg206540}. They compared parzen windows, linear SVM, linear lasso, and gradient boosting trees. They found that gradient boosting trees with pre-processed features was the most performant method for hypoxemia predictions. In their evaluation of their machine learning technique, they found that their method made more accurate hypoxemia predictions than five practicing anesthesiologists.  Lundberg et. al. showed that gradient boosting trees are very powerful methods for prediction (particularly for encoding interactions), but the preprocessed time series features used -- exponential moving averages/variances -- can be improved.  Representation learning would take advantage of the recurrence/memory that neural networks can encode, and pass latent representations into a gradient boosting tree to further improve on doctor performance.  In general, representation learning has already achieved great success in speech recognition, signal processing, object recognition, and natural language processing  \citep{DBLP:journals/corr/abs-1206-5538}. It is well suited to time series biomedical data because inherent/latent structure is a reasonable assumption in physiological data. We explore this approach and present a framework for forecasting biomedical time series data.

\section{Methods}
\textbf{Long short term memory networks}
Long short term memory networks are a sophisticated way to explicitly retain memory \citep{Hochreiter:1997:LSM:1246443.1246450}.  In comparison to the autoregressive methods used in the previous exploration of hypoxemia (Lundberg et. al.), recurrent neural networks like LSTM networks are capable of capturing more complex dependencies.  We train these networks in Python using Keras, an open source neural network library, with a Tensorflow backend \citep{chollet2015,DBLP:journals/corr/AbadiABBCCCDDDG16}.  We utilize 72 CPUs (Intel(R) Xeon(R) CPU E5-2699 v3 @ 2.30GHz) to train our networks and our tree models.

In terms of the design, the networks in this paper simply utilize two layers, because adding too many layers made convergence difficult for our application.  We found that important steps in training LSTM networks for operating room data are to impute missing values by the training mean, standardize data, and to randomize sample ordering (for time series in particular) prior to training.  To prevent overfitting, we utilized dropouts between layers as well as recurrent dropouts for the LSTM nodes  \citep{Srivastava:2014:DSW:2627435.2670313}. Using a learning rate of 0.001, rmsprop for an optimizer, and a sigmoid output layer gave us the best final results. The LSTM models were run until their validation accuracy did not improve for twenty rounds.

To the best of our knowledge this is one of the first applications of RNNs within an operating room setting, and we are unaware of any others. RNNs have been applied in other health settings: Lipton et al., (2015) applied LSTM networks in a clinical setting, Che et al. applied GRUs for imputation in clinical and synthetic data sets (2016) \citep{DBLP:journals/corr/ChePCSL16,DBLP:journals/corr/LiptonKEW15}. Chauhan and Vig (2015) and Rajpurkar et. al. (2017) have done work that focused on univariate applications of neural networks in health settings \citep{cv15,DBLP:journals/corr/RajpurkarHHBN17}.

\textbf{Gradient Boosting Trees}
Gradient boosting trees work well in practice due to their ease of use and flexibility.  Imputing, standardizing, and randomizing are all unnecessary because gradient boosting trees are based on splits in the training data.  We postulate that gradient boosting trees are better at incorporating important static features in predictions than LSTM networks and saw very good performance with simple methods of processing time series features (exponential moving averages/variances).  We found that a learning rate of 0.02, a max tree depth of 6, subsampling rate of 0.5, and a logistic objective gave us good performance.  All XGB models were run until their validation accuracy was non-improving for five rounds.  We train these models in Python using xgboost, an open source library for gradient boosting trees \citep{Chen:2016:XST:2939672.2939785}.

\textbf{Methodology}
\begin{enumerate}[topsep=0pt,itemsep=-1ex,partopsep=1ex,parsep=1ex]
\item Run XGB using processed times series data (EMA/EMV) and raw static data.
\item Identify the most important features.
\item Run supervised learning with univariate LSTM networks on these features.
\item Use second to last LSTM layer to create features for XGB.
\item Retrain XGB model with additional hidden features for the final model.
\end{enumerate}

\section{Experimental Results}
\label{results}

\textbf{Data} 57,000 surgeries containing real-time features sampled minute by minute such as SaO2 (blood oxygen), ETCO2 (exhaled carbon dioxide), etc. as well as static summary information such as height, weight, ASA codes, etc. obtained under appropriate Institutional Review Board (IRB) approval.  After splitting surgeries into multiple time points, there are $\approx 8,000,000$ samples with $\approx 120,000$ positive examples.  These labels represent a time series binary hypoxemia classification problem where SaO2 less than 92\% is considered hypoxemia.

\textbf{Evaluation Metric}  \emph{Area under the precision-recall (PR) curve} is our evaluation metric. PR curves are widely used for binary classification tasks to summarize the predictive accuracy of a model. Rather than ROC curves, PR curves are better suited to classification problems with imbalanced labels. True positives ($TP$) are positive sample points that are classified as positive whereas true negatives ($TN$) are negative sample points that are classified as negative.  Then, false positives ($FP$) are negative sample points that are classified as positive whereas false negatives ($FN$) are positive sample points that are classified as negative.  Precision is defined as $\frac{tp}{tp+fp}$ and recall is $\frac{tp}{tp+fn}$. The PR curve is plotted with precision (y-axis) for each value of recall (x-axis). In order to summarize this curve, it is conventional to use area under the curve (AUC) to measure prediction performance. 

\textbf{Performance}
We demonstrate that long short term memory networks outperform autoregressive methods.

\begin{figure}[ht]
\includegraphics[width=0.5\textwidth]{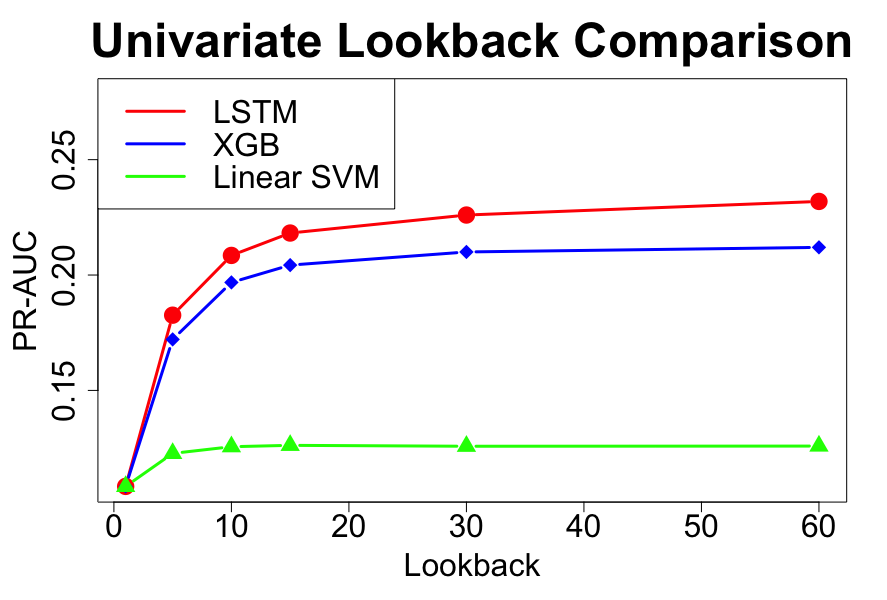}
\centering
  \caption{Univariate models trained on SaO2 only (the feature that determines hypoxemia) with different lookbacks.  All methods attempt to predict whether there will be desaturation in the next five minutes.}
  \label{autoregressive}
\end{figure}

Figure \ref{autoregressive} reveals two things.  First, we see that on a univariate signal, LSTM networks substantially outperform autoregressive methods -- even performant ones like gradient boosting trees. Second, we see that LSTM networks improve in performance even after the autoregressive methods saturate in terms of accuracy.  Utilizing lookbacks of 60 rather than 30 markedly improves LSTM network performance -- suggesting that they successfully identify complex long term patterns relevant to prediction.  Moving forward, the LSTM networks use lookbacks of 60 minutes to predict desaturation.  

\begin{table}[ht]
  \caption{LSTM network Performance on Hypoxemia}
  \label{lstm-table}
  \centering
  \begin{tabular}{llll}
    \toprule
    & Model & Test PR-AUC & Train Time\\
    \midrule
    1 & LSTM network (200x200) on raw (SaO2) & 0.23183 & $\approx 43.00$ hours\\
    2 & LSTM network (400x400) on raw (40 features) & 0.23315 & $\approx 62.67$ hours\\
    \bottomrule
  \end{tabular}
  \caption*{PR-AUC on the test set.  We also evaluate the time required to train the neural network.  Here (200x200) denotes that there were two layers each with 200 LSTM nodes.  The raw (SaO2) denotes the input to the LSTM network.}
\end{table}

\begin{figure}[ht!]
\includegraphics[width=0.5\textwidth]{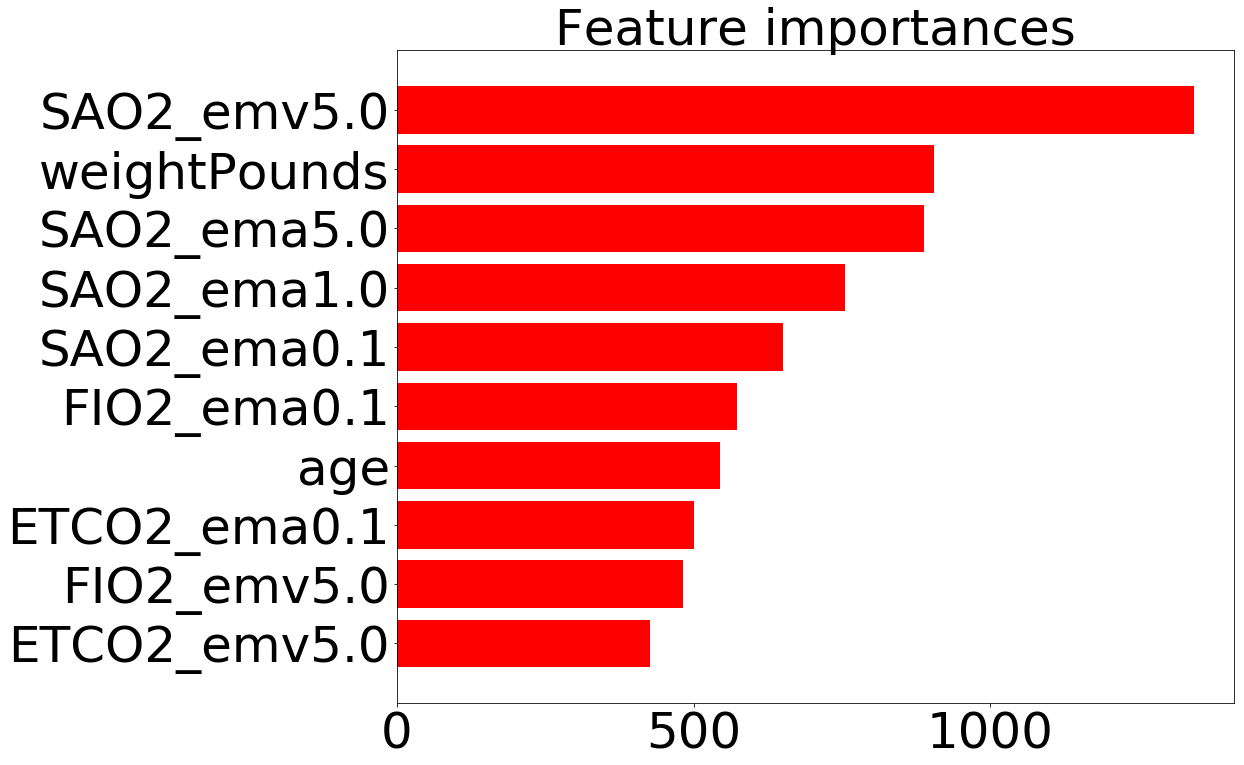}
\centering
  \caption{Feature attribution for Table \ref{xgb-table} Model 2.}
  \label{feature_importances}
\end{figure}

Although multivariate LSTM networks to predict hypoxemia appear promising, they take a substantial amount of time to train.  Despite having twice as many nodes, Table \ref{lstm-table} Model 2 (a multivariate LSTM network with 40 features) only gives a slight improvement over Table \ref{lstm-table} Model 1 (a univariate LSTM network with SaO2).  Based on training the gradient boosting model, we suspect that static features contain more information relevant to prediction than the multivariate LSTM networks are capable of utilizing (Figure \ref{feature_importances} justifies the importance of static features).  Additionally, the cost of training the LSTM network is already increasing drastically with the model size.  In order to successfully capture all relationships in a complete multivariate network, the network size would need to increase to an intractably large size.  Preliminary results with a GeForce GTX 1050 GPU were approximately three times faster, but a multivariate network would likely still be intractable.

\begin{table}[ht!]
  \caption{XGB Performance on Hypoxemia}
  \label{xgb-table}
  \centering
  \begin{tabular}{lll}
    \toprule
     & Model & Test PR-AUC\\
    \midrule
    1 & XGB on processed (SaO2 only) & 0.20580\\
    2 & XGB on processed (40 features) & 0.22999\\
    3 & XGB on LSTM hidden (200 features) & 0.23354\\
    4 & XGB on processed (40 features) + LSTM output (1 feature) & 0.24714\\
    5 & XGB on processed (40 features) + LSTM hidden (200 features) & 0.24836\\
    6 & XGB on processed - no SaO2 (39 features) + LSTM output (1 feature) & 0.24251\\
    7 & XGB on processed - no SaO2 (39 features) + LSTM hidden (200 features) & 0.24678\\
    \bottomrule
  \end{tabular}
  \caption*{PR-AUC on the test set.  We don't report the time to train the gradient boosting trees because the training times were fairly short -- with the slowest model taking approximately 4 hours to train.  Processed denotes three exponential moving averages (EMA) with $\alpha=5.0, 1.0,$ and $0.1$ as well as an exponential moving variance (EMV) with $\alpha=5.0$ for all time series features.  LSTM output denotes using the probability for hypoxemia output by the final sigmoid layer as an input to XGB.  LSTM hidden denotes using the outputs from the penultimate LSTM layer as inputs to XGB. The LSTM output/hidden features are generating using Model 1 from Table \ref{lstm-table}.}
\end{table}

Then, Table \ref{xgb-table} shows that XGB utilizing only processed SaO2 (Table \ref{xgb-table} Model 1) does not immediately do well.  Yet when XGB has access to all 40 processed features (Table \ref{xgb-table} Model 2) PR-AUC increases drastically -- from $0.20580$ to $0.22999$. This result suggests that gradient boosting trees are able to leverage the 39 remaining features to improve predictive accuracy in a way that LSTM networks alone could not.  In figure \ref{feature_importances}, SaO2 dominates the top 10 most important features for hypoxemia prediction, but static features like weight and age are also of importance.  Training models to effectively utilize these additional features did not appear tractable with LSTM networks (Table \ref{lstm-table}), thus motivating our exploration of a hybrid approach.

Then, in Model 3 of Table \ref{xgb-table} XGB utilizing LSTM latent representations gives slightly better performance than the LSTM network alone, suggesting that gradient boosting trees are able to capture the non-linearity in the final layer of the LSTM network. Next, Model 4 shows that augmenting XGB with a supervised latent representation of SaO2 from our LSTM network gives substantially better performance than either individual model.  Since gradient boosting trees work well on hidden features alone (Model 3), we combine them in Model 5 with the processed features, which yields the most performant model for hypoxemia prediction.  Because we included the EMAs/EMV in Models 4 and 5, how much of the predictive performance is due to the EMAs/EMV and how much is due to the hidden LSTM features is still an open question.  In Models 6 and 7 the hidden features are revealed to be much more informative than the single predictive probability, suggesting that the hidden features not only serve as fairly powerful surrogates for the EMAs/EMV, but that they also capture additional information that EMAs/EMV features do not.  This suggests that supervised representation learning is capable of finding sensible representations.  If these representations can be learned for many tasks and data sets, machine learning research scientists may be able to work additively on impactful questions in health.  Through representation learning, motifs in time series data can be conveyed between research groups with access to different data sets in anonymous yet meaningful ways -- protecting patient privacy.

At a high level, this hybrid approach is very simplistic.  Yet this simplicity offers two nice properties, the first being generalizability -- our hybrid approach to machine learning can easily be applied to any problem that would benefit from more sophisticated time series processing. The second property being accessibility -- using the open source packages available, these two methodologies can easily be combined in very performative ways for biomedical data. Accessibility ideally encourages a trend towards representation learning as a means to make additive progress (collaboration rather than competition) on biomedical prediction tasks.

\bibliographystyle{unsrt}


\end{document}